# Rational Choice and Artificial Intelligence

Tshilidzi Marwala
University of Johannesburg
tmarwala@gmail.com

**Summary.** The theory of rational choice assumes that when people make decisions they do so in order to maximize their utility. In order to achieve this goal they ought to use all the information available and consider all the choices available to choose an optimal choice. This paper investigates what happens when decisions are made by artificially intelligent machines in the market rather than human beings. Firstly, the expectations of the future are more consistent if they are made by an artificially intelligent machine and the decisions are more rational and thus marketplace becomes more rational.

## Introduction

This paper is basically on rational decision making, through the theory of rational choice, and this is a complex process and has perplexed thinkers for a very long time. On making a decision rationally a person should take into account that other players may be acting irrationally. In this situation, the rational decision that is taken will factor into account the irrational reaction from the other players. Suppose a doctor prescribes medicine to a superstitious individual. That decision making process (rational) of prescribing medicine might take into account of the individual's superstitious tendencies (irrational) and thus also prescribes that the patient should be supervised on taking the medication. So in essence the decision taken has factored into account of the fact that the patient will act irrationally. Rational choice is a mechanism of choosing using complete information and evaluating all options to select an option that globally maximizes untility.

In prehistoric society, decision making was jumbled with superstitions (Marwala, 2014). For example, there is an old superstitious belief that if a person encounters a black cat crossing the road then that person will encounter bad luck. The other superstition among the Venda people in South Africa is that if the owl crows in your house then there will be death in the family. How these superstitions came about is not known but one can speculate that perhaps sometimes in the past a powerful person in the community, perhaps the King, by chance encountered a black cat crossing the road and he had bad luck subsequently. Whatever the explanations for these superstitions may be, the fact remains that superstitions have been and remain part of who we are as human beings in this present state of our evolution. Perhaps sometimes in the future we will evolve into some other species which is not superstitious and the fourth industrial revolution or man-machine system offers us a unique opportunity for that to happen.

Superstition can be viewed as supernatural causality where something is caused by another without them being connected to one another. The idea of one event causing another without any connection between them whatsoever is irrational. Making decisions based on superstitious assumptions is irrational decision making and is the antithesis of this paper which is on rational decision choice.

This paper basically proposes that the basic mechanisms of rational decision making are causality and correlation machines to create rational expectations machines and optimization select appropriate option. Suppose we compare two ways of moving from A to B. The first choice is to use a car and the second one is to use a bicycle. The rational way is not just to select the choice which minimizes time but to compare this with the cost of the action. Rational decision making is a process of reaching a decision that maximizes utility and these decisions are arrived at based on relevant information and by applying sound logic and optimizing resources (Marwala, 2015).

Nobel Prize Laureate Herbert Simon realized that on making a rational decision one does not always have all the information and the logic that is used is far from perfect and, consequently, he introduced the concept of bounded rationality (Simon, 1991). Marwala (2014) in a book that extends the theory of bounded rationality developed this notion that with the advent of artificial intelligence the bounds of rationality that Herbert Simon's theory prescribes are in fact flexible. The next section explores what rational decision making is and why it is important.

**What is Rational Choice?**

It is important to first understand what rational choice is and to do this it is important to understand the meaning of the words rational and choice (Green and Shapiro, 1994; Friedman, 1996). According to the google dictionary, rational is defined as "based on or in accordance with reason or logic" while choice is defined as "the act of choosing between two or more possibilities'. Rational choice is a process of making decisions based on relevant information, in a logical, timely and optimized manner. Suppose a man called Thendo wants to make a decision on how much coffee he should drink today and he calls his sister Denga to find out the color of shoes she is wearing and uses this information to decide on the course of action on how much coffee he will drink on that day. This will be an irrational decision making process because Thendo is using irrelevant information (the color of shoes his sister Denga is wearing) to decide how much coffee he will drink. If on the same token Thendo decides that every time he takes a sip of that coffee he needs to walk for 1km, we then will conclude that he is acting irrationally because he is wasting energy unnecessarily by walking 1km in order to take a sip of coffee. This unnecessary wastage of energy is irrational and in mathematical terminology we describe this as an un-optimized solution. If Thendo decides that every time he sips the coffee he first pours it from one glass to another for no other reason except to fulfil the task of drinking coffee, then he is using an irrational course of action because this is not only illogical but also un-optimized.

An illustration of a rational decision making framework is shown in Figure 1. This figure shows that on undertaking a rational decision making process one studies the environment or more technically a decision making space. Then the next step is to identify relevant information necessary for decision making. Then this information is presented to a decision engine which is logical and consistent and evaluates all possibilities and their respective utilities and then selects the decision that offers the highest utility. Any weakness in this framework such as absence of all information, or information that is imprecice and imperfect or not being able to evaluate all possibilities limits the theory of rational choice and this then becomes a bounded rational choice proposed by Herbert Simon (1991). What is termed classical economics is based on the assumption that the agents in economics makes decisions based on the theory of rational choice. The fact that this agent is a human being has led many researchers to study how human beings make decisions and this is now what is called behavioral economics. Kahneman in his book *Thinking Fast and*

*Slow* explores behavioral economics extensively (Kahneman, 1991). Today decisions are more and more made by artificially intelligent machines. These artificial intelligent machines bring several aspects that make the assumption of rationality to be stronger than when decisions are made by human beings. In this regard the machine made decisions are not as irrational as when they are made by human beings, however, they are not fully rational and therefore are still subjected to the Herbert Simon's theory of bounded rationality.

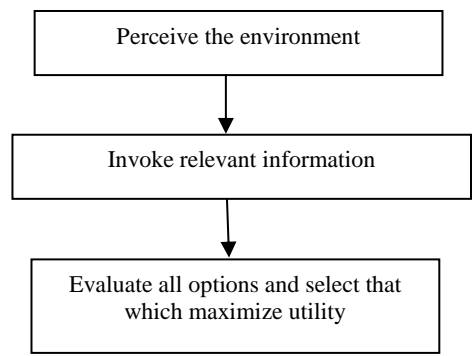

Figure 1 Steps for rational decision making

For example, if in some isolated economic system all the decision making agents are human beings, what will be the characteristics of that economy? That economy will be characterized by behavioral economics. If however, all the decision making agents are artificial intelligent machines that make decisions based on rational choice, then it probably will be characterized by neoclassical economics. If half of these decision making agents are humans and the others are machines then some economic system which is half neoclassical and half behavioral will emerge.

**Information**

The theory of rational choice is based on the fact that the information used to make decisions is complete and precise. This, however, is not physically realistic. The word information is from the verb informs and it means that which informs (Casagrande, 1999). One of the ways in which information is represented is data which need to be sensed (Dusenbery, 1992; Vigo, 2013). For example, if one requires the information on the temperature of a body one can use a thermometer to achieve this goal. The information from the thermometer is data but it is really the volume of mecury which is not a representation of temperature but is correlated to temperature within certain conditions. For example, the same thermometer cannot be used for measuring very high temperatures (Vigo, 2014). Measuring or acquiring information is not a precise science and is subject to uncertainties. These uncertainties are due to the fact that by measuring a metric from an object the act of measuring changes the object and therefore the results are no longer the intended results. In Physics a German Nobel Laureate Werner Heisenberg proposed the uncertainty principle stating that one cannot measure the position and the speed of an electron accurately at the same time (Heisenberg, 1927; Sen, 2014). If one knows the speed one does not know its position. On the other hand if one knows its position then one does not know its speed accurately. It turns out that this paradox is not because the measurement interferes with the speed or position but it is just the fundamental nature of quantum mechanics. Getting complete and precise information is not possibile so we have to deal with uncertainty, incompleteness and inaccuracy and this has impact on the quality of decisions that are made.

**Choices**

The other factor in the theory of rational choice is the act of choosing options from many choices available (Hsee et. al., 1999; Irons and Hepburn, 2007). Suppose one wants to commercially fly from Johannesburg to New York. There are many choices to consider. For example, one can fly from Johannesburg straight to New York or Johannesburg, Dubai then New York or Johannesburg, London then New York. The options available are thousands. In fact if human beings have to make these choices they cannot possibly be able to consider all these options in a reasonable time without the use of some software such as an artificial intelligence optimization techniques (Schwartz, 2005; Reed et. al., 1999). Suppose it is raining outside and one needs to walk to the gate to pick up some mail. Suppose there is an umbrella and therefore are only two choices i.e. use an umbrella and walk to the gate and pick up some mail or do the same without an umbrella. The one that maximizes comfort i.e. utility is to pick up an umbrella and walk to the gate. This is a simple example because there are only two choices. In many instances there are thousands of viable options and searching through all of them requires advanced search algorithms such as genetic algorithm. Furthermore, there might even be infinite choices making the prospect of a viable choice not reachable. In fact that one cannot be able to explore all options is more of a norm than an exception.

**Optimization**

On our task to identify a suitable route to travel to New York from Johannesburg described above we can frame a problem in such a way that we want to identify a route which is the shortest or in which we shall spend the least amount of time on air. This is an optimization problem with an objective function of finding a route which minimizes the time on air (Dixit, 1990; Rotemberg and Woodford, 1997). The process of identifying such route is called optimization and in this case it is to fly directly from Johannesburg to New York. This objective function is also called a cost function and is that function which a person wants to minimize (Diewert, 2008). We use our understanding of topology i.e. intelligence to identify that the shortest distance is a straight line and to infer that a direct route is the shortest one. Artificial intelligence has been used widely to solve such problems (Battiti et al., 2008)).

**Rational Choice**

The theory of rational choice state that given several options a rational human being will choose that option that maximizes his utility (Becker, 1976; Hedström and Stern, 2008; Lohmann, 2008; Grüne-Yanoff, 2012). This theory assumes several factors and these include the fact that the decision maker has all the information that he can use to generate several options. In addition, that he can be able to construct all options as well as calculate the resulting utility for each option. An illustration of this concept is shown in Figure 1. There are a number of principles that govern rational chice and these are completeness and transitivity. Completeness means that all pairs of alternative options can be compared to one another whereas transitivity means if option 1 is better than option 2 which is in turn better than option 3 then option 1 is better than option 3. In addition there are different classes of preferences and these are strict, weak and indeference. Strict preference is when one option is preferred over another whereas indeference is when a person does not prefer one option over another. This Figure shows a problem with $n$ options and their corresponding utility. In example of identifying a shortest time to fly from Johannesburg to New York the utility function is the inverse of the the time it takes for this trip. In this regard, the lower the time it takes the higher it takes to reach New York. In this regard supppose Johannesburg-New York takes 18 hours, Johannesburg-Dubai-New York 36 hours, Johannesbur-London-New York takes 24 hours, Johannesburg-Paris-New York 26 hours and so on. A table can be generated that takes this trips into utility and this is indicated in Table 1. The uility is calculated as an inverse of the time it takes i.e. one divided by the time it takes. This example, shows that to maximize utility, then one ought to fly directly from Johannesburg to New York.

Table 1 An illustration of the calculation of utility

| Options | Time (hours) | Utility |
|---|---|---|
| JHB-NY | 18 | 0.05555556 |
| JHB-DB-NY | 36 | 0.02777778 |
| JHB-LN-NY | 24 | 0.04166667 |
| JHB-PR-NY | 26 | 0.03846154 |

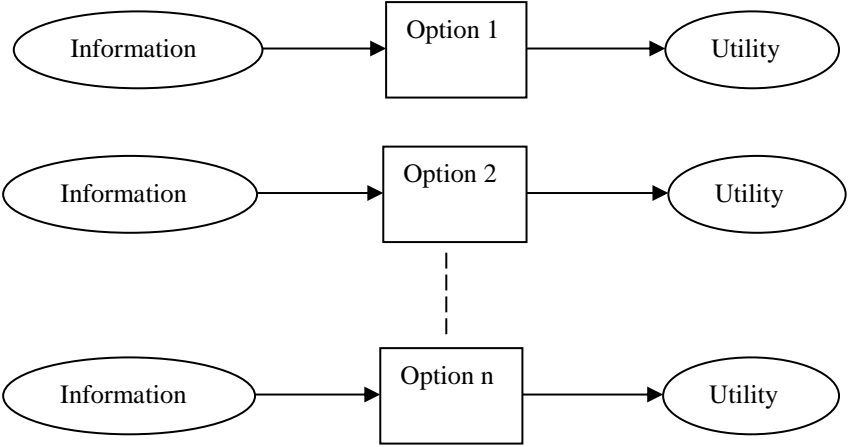

Figure 1 An illustration of several options and their respective utilities

Figure 1 can be simplified into Figure 2 where there is a generalized options generator that takes information and comes up with the optimal utility. An example of this is when options can be represented in the form of a function which takes information and predict alternative utilities given certain conditions and then identify through some optimization routine the option that maximizes utility.

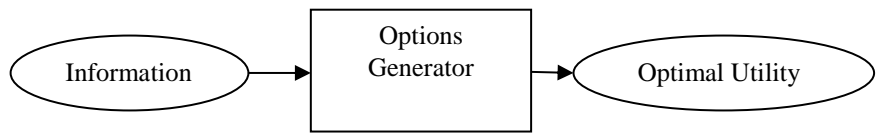

Figure 2 A generalized options generator that takes information and gives with the optimal utility

**Rational Choice and Opportunity Cost**

The fact that rational choice entails evaluatig alternatives invokes the cooncept of opportunity cost. The concept of opportunity cost was introduced by von Wieser (1914). According investopedia: "Opportunity cost refers to a benefit that a person could have received, but gave up, to take another course of action" (Anonymous, 2017). When a person is presented with several options as shown in Figure 1, a rational person will choose the option that maximizes her utility. The second best option for a rational person is that with a utility that is equal to (or more than the one that was chosen-this is not possible because he has already chosen the optimal choice. It will not be rational for a person to choose an option which has a lower utility than the option chosen. Perhaps there is a need to intoduce the rational opportunity cost which is the opportunity forgone that gives the expected utility equal to that which a rational person has pursued (a rational person maximizes utility). This is possible as specially when there is more than one optimal solution.

**Rational Choice and Artificial Intelligence**

On applying the theory of rational choice a mechanism of generatng options available needs to be established. For example, in the case of a trip from Johanesburg to New York, the options available are generated by connecting all the routes from Johannesburg to New York with one stopover, then 2, then three etc. The options available are numerous. For example, there is a problem of inflation targeting that has been studied extensively where the goal is to use economic variables including the previous interest rate to predict inflation rate and then to identify the interest rate that gives the desired inflation (Marwala, 2012). The model which takes these economic variables and mapping these to the inflation rate can be achieved using neural networks and any other learning machine (Marwala and Lagazio, 2011).

Neural networks is based on the functioning of the brain and therefore has much of the attributes of the brain such as the neuron in the brain being the processing element in neural networks, dendrites being the combining function, cell body being the transfer function, axon being the element output and synapses being the weights. Neural networks is a powerful machine that has been used to model complex sytems (Marwala, 2007), estimate missing data (Abdella and Marwala, 2005; Marwala, 2009), predict electrical faults (Marwala, 2012), economic modeling (Marwala, 2013), rational decision making (Marwala, 2014), understanding causality (2015) and model mechanical structures (Marwala et. al., 2006; Marwala et al., 2017). These learning machines have free parameters and acivation functions and the free parameters are estimated from the data using some optimization method. In this way, whenever economic variables are given then interest rate can be estimated.

The idea of predicting interest rate based the previous interest rates is called the theory of adaptive expectations (Galbács, 2015). The theory of adaptive expectations basically states that we can predict the future of an object based on its past. Unfortunately the theory of adaptive expectations has been found to result in systematic errors due to bias. The concept of including other variables in addition to the previous interest rate to predict the future interest rate is called the theory of rational expectations and this eliminates systematic bias and errors (Muth, 1971; Sargent, 1987; Savin, 1987).

After identifying a model that predicts the future interest rates based on the theory of rational expectations which is made more valid by incorporating the use of artificial intelligence, data fusion, big data and text analysis then one can identify the interest rate that will maximize the attainment of the desired inflation rate using an optimization routine. Optimization methods that are able to identify global optimum solutions are normally based on evolutionary techniques such as genetic algorithm, particle swarm optimization and simulated annealing (Perez and Marwala, 2008). Genetic algorithm is a stochastic optimization method which is based on the principles of evolution where mutation, crossover and reproduction are used to identify an optimal solution (Crossingham and Marwala, 2008). Particle swarm optimization is a stochastic optimization method which is based on the principles of local and group intelligence as observed in the swarm of birds when they are identifying a roost and this algorithm can be used to identify an optimal solution (Mthembu et. al., 2008). These optimization method can be used within the context of the theory of rational choice to identify a course of action that maximizes utility.

The implication of artificial intelligence on the theory of rational choice are as follows: (1) Artificial intelligence makes use of information in various format e.g. text, pictures, internet and fuse them to make better rational expectations of the future bringing decision making closer to rational choice; (2) On making the optimal choice global optimization methods such as genetic algorithm gives higher probability of identifying a global optimum utility thereby bringing decision making closer to the theory of rational choice.

**Interstate Conflict and Rational Choice**

Interstate conflict is a major source of economic instability and the rational course of action for rational states is to minimize it or maximize the attainment of peace. This is achieved by creating a model which takes several input variables and predict the risk of conflict. This prediction of the risk of conflict is done by assuming the theory of rational expectations using neural networks to predict the conflict outcome (zero for peace and one for conflict). This outcome is deemed to be the utility. Then using this neural networks, controllable variables are identified and tuned using genetic algorithm optimization method to produce a peaceful outcome as was done by Marwala and Lagazio (2011). The neural network is trained using seven dyadic independent variables and these are *Allies*, a binary measure coded 1 if the members of a dyad are linked by any form of military alliance, *Contingency* which is binary, coded 1 if both states are geographically contiguous, *Distance* which is an interval measure of the distance between the two states' capitals, *Major power* which is a binary variable coded 1 if either or both states in the dyad are a major power, *Democracy* which is measured on a scale where 10 is an extreme democracy and -10 is an extreme autocracy, *Dependency* which is measured as the sum of the countries import and export with its partner divided by the Gross Domestic Product of the stronger country and *Capability* which is the logarithm, to the base 10, of the ratio of the total population plus the number of people in urban areas plus industrial energy consumption plus iron and steel production plus the number of military personnel in active duty plus military expenditure in dollars in the last 5 years measured on stronger country to weak country. We lag all independent variables by one year to make temporally plausible any inference of causation. Of the 7 dyadic variables used in this chapter, there are only 4 that are controllable and these are: *Democracy*, *Allies*, *Capability* and *Dependency*.

As was done by Marwala and Lagazio (2011), the Golden section search (GSS) technique was used for the single strategy approach and simulated annealing was used for multiple strategy approach. This technique is shown in Figure 3.

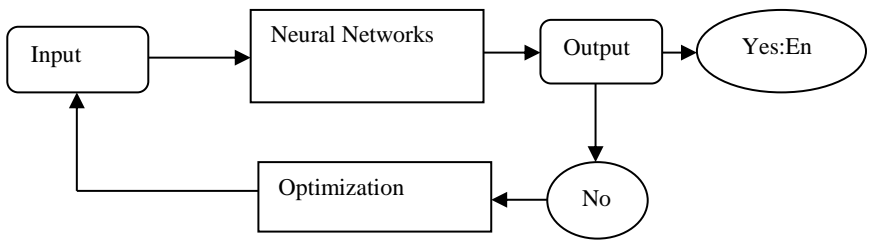

**Figure 3** Feedback control loop that uses Bayesian neural networks and an optimization method

When the control strategies were implemented, the results shown in Figure 4 were obtained (Marwala and Lagazio, 2011). These results show that, for a single strategy approach, where *Democracy* is the controlling dyadic variable, 90% of the 286 conflicts could have been avoided. When the controlling the dyadic variable *Allies* as the only variable used to bring about peace, it was found that 77% of the 286 conflicts could have been avoided. When either *Dependency* or *Capability* was used as a single controlling variable, 98% and 99% of the 286 conflicts could have been avoided, respectively. In relation to the multiple strategy approach, when all the controllable variables were used simultaneously to bring about peace, all the 286 conflicts could have been avoided.

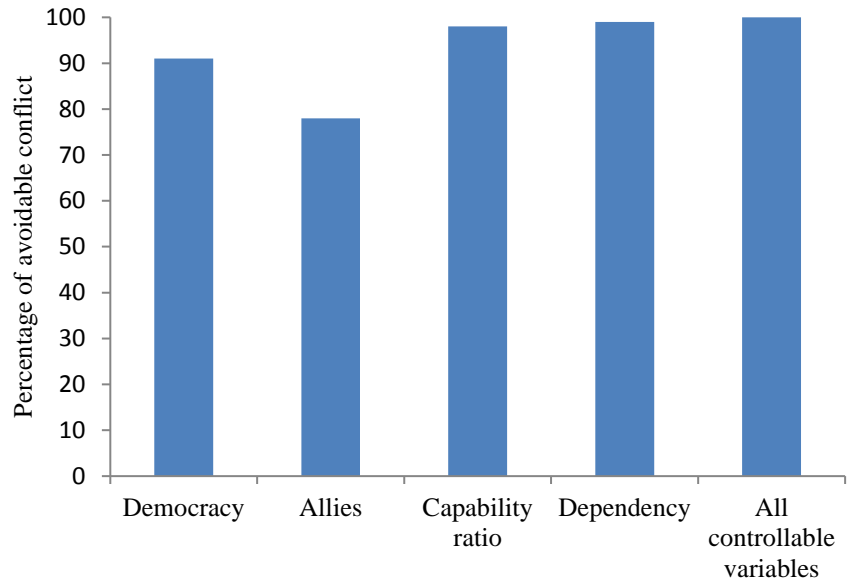

**Figure 4** Graph showing the proportion of past conflicts that could have been avoided

These results demonstrate that all the variables need to change less compared to the single approach to produce peace. Nevertheless, the change required for *Democracy* and *Dependency* is only marginally different from the one required by these variables in the single strategy, although *Allies* and *Capability* require to change considerably less compared with the single approach. This indicates the significance of democracy and dependency to achieve peace. From these results, it is evident that policy makers should focus on democratization and economic cooperation to maximize attainment of peace. They should resolve which method, single or multiple, is the most suitable on the basis of how easy and cost efficient an intervention with the identified controllable variables is, and how long it would take to control them.

**Conclusion**

This chapter introduced rational choice and how it is changed through the advances of artificial intelligence. Furthermore, it introduced how the theory of rational choice changes the definination of the concept of opportunity costs. It proposes a frramework of rational choice which is based on the theory of rational expectations to create a causal model between the input and output using neural netwoks and optimization to change input to obtain the desired output. The theory of rational choice and artificial intelligence are used to maximize the attainment of peace.

**References**

Abdella, M. and Marwala, T. (2005) Treatment of missing data using neural networks and genetic algorithms. IEEE International Joint Neural Networks, Montreal, pp. 598-603.

Anonymous (2017) "Opportunity Cost". Investopedia. Last Retrieved 2017-03-18.

Battiti, R., Brunato, M. and Mascia, F. (2008). Reactive Search and Intelligent Optimization. Springer Verlag.

Becker, G.S. (1976). The Economic Approach to Human Behavior. Chicago. Description and scroll to chapter-preview links.

Casagrande, D. (1999). "Information as verb: Re-conceptualizing information for cognitive and ecological models". Journal of Ecological Anthropology. 3 (1): 4–13. doi:10.5038/2162-4593.3.1.1.

Crossingham, B. and Marwala, T. (2008), Using Genetic Algorithms to Optimise Rough Set Partition Sizes for HIV Data Analysis. Advances in Intelligent and Distributed Computing, Studies in Computational Intelligence, Volume 78, pp. 245-250

Diewert, W.E. (2008). "Cost functions," The New Palgrave Dictionary of Economics, 2nd Edition Contents.

Dixit, A.K. (1990). Optimization in Economic Theory, 2nd ed., Oxford. Description and contents preview.

Dusenbery, D.B. (1992). Sensory Ecology. W.H. Freeman., New York.

Friedman, J. (1996). The Rational Choice Controversy. Yale University Press.

Galbács, P. (2015). The Theory of New Classical Macroeconomics. A Positive Critique. Heidelberg/New York/Dordrecht/London: Springer.

Green, D.P. and Shapiro, I. (1994). Pathologies of Rational Choice Theory: A Critique of Applications in Political Science. Yale University Press.

Grüne-Yanoff, T. (2012). "Paradoxes of Rational Choice Theory". In Sabine Roeser, Rafaela Hillerbrand, Per Sandin, Martin Peterson. Handbook of Risk Theory. pp. 499–516. doi:10.1007/978-94-007-1433-5_19.

Hedström, P. and Stern, C. (2008). "rational choice and sociology," The New Palgrave Dictionary of Economics, 2nd Edition. Abstract.

Heisenberg, W. (1927), "Über den anschaulichen Inhalt der quantentheoretischen Kinematik und Mechanik", Zeitschrift für Physik (in German), 43 (3–4): 172–198

Hsee, C.K., Loewenstein, G.F., Blount, S., Bazerman, M.H. (1999). Preference reversals between joint and separate evaluations of option: A review and theoretical analysis. Psychological Bulletin 125(5), 576–590.

Irons, B. and C. Hepburn. 2007. "Regret Theory and the Tyranny of Choice." The Economic Record. 83(261): 191–203.

Kahneman, D. (2011). Thinking, Fast and Slow. Macmillan.

Marwala, T. (2015). Causality, Correlation, and Artificial Intelligence for Rational Decision Making. Singapore: World Scientific.

Lohmann, S. (2008). "Rational choice and political science,"The New Palgrave Dictionary of Economics, 2nd Edition.Abstract.

Marwala, T. (2015). Causality, Correlation, and Artificial Intelligence for Rational Decision Making. Singapore: World Scientific. ISBN 978-9-814-63086-3.

Marwala, T. (2014). Artificial Intelligence Techniques for Rational Decision Making. Heidelberg: Springer.

Marwala, T. (2013). Economic Modeling Using Artificial Intelligence Methods. Heidelberg: Springer.

Marwala, T. (2012). Condition Monitoring Using Computational Intelligence Methods. Heidelberg: Springer.

Marwala, T. (2010). Finite Element Model Updating Using Computational Intelligence Techniques: Applications to Structural Dynamics. Heidelberg: Springer.

Marwala, T. (2009). Computational Intelligence for Missing Data Imputation, Estimation, and Management: Knowledge Optimization Techniques. Pennsylvania: IGI Global. ISBN 978-1-60566-336-4.

Marwala, T. (2007). Computational Intelligence for Modelling Complex Systems. Delhi: Research India Publications.

Marwala, T., Boulkaibet, I, and Adhikari S. Probabilistic Finite Element Model Updating Using Bayesian Statistics: Applications to Aeronautical and Mechanical Engineering. John Wiley and Sons, 2016, ISBN: 978-1-119-15303-0.

Marwala, T. and Lagazio, M. (2011). Militarized Conflict Modeling Using Computational Intelligence. Heidelberg: Springer. ISBN 978-0-85729-789-1.

Marwala, T., Mahola, U. and Nelwamondo, F.V. (2006) Hidden Markov models and Gaussian mixture models for bearing fault detection using fractals. International Joint Conference on Neural Networks, pp. 3237-3242

Mthembu, L., Marwala, T., Friswell, M.I. and Adhikari, S. (2011), Finite Element Model Selection Using Particle Swarm Optimization. Conference Proceedings of the Society for Experimental Mechanics Series, 1, Volume 13, Dynamics of Civil Structures, Volume 4, Springer London, pp. 41-52, Tom Proulx (Editor)

Muth, J.F. (1961) "Rational Expectations and the Theory of Price Movements" reprinted in The new classical macroeconomics. Volume 1. (1992): 3–23 (International Library of Critical Writings in Economics, vol. 19. Aldershot, UK: Elgar.)

Perez, M. and Marwala, T. (2008) Stochastic optimization approaches for solving Sudoku. arXiv preprint arXiv:0805.0697

Reed, D.D., DiGennaro Reed, F.D., Chok, J. and Brozyna, G.A. (2011). The 'tyranny of choice': Choice overload as a possible instance of effort discounting. The Psychological Record, 61(4), 547-60.

Rotemberg, J. and Woodford, M. (1997), "An Optimization-based Econometric Framework for the Evaluation of Monetary Policy" NBER Macroeconomics Annual, Vol. 12, pp. 297-346.

Sargent, T.J. (1987). "Rational expectations," The New Palgrave: A Dictionary of Economics, v. 4, pp. 76–79.

Savin, N.E. (1987). "Rational expectations: econometric implications," The New Palgrave: A Dictionary of Economics, v. 4, pp. 79–85.

Schwartz, B. (2005). The Paradox of Choice: why more is less. Harper Perennial.

Sen, D. (2014). "The uncertainty relations in quantum mechanics" (PDF). Current Science. 107 (2): 203–218.

Simon, H. (1991). "Bounded Rationality and Organizational Learning". Organization Science. 2 (1): 125–134.

Vigo, R. (2011). "Representational information: a new general notion and measure of information". Information Sciences. 181: 4847–4859. doi:10.1016/j.ins.2011.05.020.

Vigo, R (2013). "Complexity over Uncertainty in Generalized Representational Information Theory (GRIT): A Structure-Sensitive General Theory of Information". Information. 4 (1): 1–30. doi:10.3390/info4010001.

Vigo, R. (2014). Mathematical Principles of Human Conceptual Behavior: The Structural Nature of Conceptual Representation and Processing. Scientific Psychology Series, Routledge, New York and London; ISBN 0415714362.

von Wieser, F. (1914). Theorie der gesellschaftlichen Wirtschaft [Theory of Social Economics] (in German). New York: Adelphi.